\documentclass[11pt, a4paper, twocolumn]{article}
\usepackage[affil-it]{authblk} 

\usepackage{amsmath}
\usepackage{listings, courier}

\usepackage{microtype}
\usepackage{graphicx}
\usepackage{subfigure}
\usepackage{booktabs} 
\usepackage{algorithm}
\usepackage{algpseudocode}

\usepackage{hyperref}

\graphicspath{ {figures/} }

\title{Diverse mini-batch Active Learning}

\author{Fedor Zhdanov\\
        fedor@amazon.com}
\affil{Amazon Research}

\begin{document}
	
\maketitle

	\begin{abstract}
	We study the problem of reducing the amount of labeled training data required to train supervised classification models. We approach it by leveraging Active Learning, through sequential selection of examples which benefit the model most. Selecting examples one by one is not practical for the amount of training examples required by the modern Deep Learning models. We consider the mini-batch Active Learning setting, where several examples are selected at once. We present an approach which takes into account both informativeness of the examples for the model, as well as the diversity of the examples in a mini-batch.
	By using the well studied K-means clustering algorithm, this approach scales better than the previously proposed approaches, and achieves comparable or better performance.
	\end{abstract}
	
	\section{Introduction}
	Supervised Machine Learning models are used in increasingly many areas to make business decisions. To train modern Deep Learning algorithms, large quantities of labeled data are required, and collecting the labels through the annotation systems such as Amazon Mechanical Turk is a costly process. Active Learning (AL) is an area which helps in reducing the amount of labeled data required to train models to the same levels of accuracy as those trained on the full dataset. Active Learning algorithms suggest which examples should be annotated first, in order for the model to gain most information about the problem.
	
	Most AL algorithms suggest examples one at a time, with the most potentially informative one suggested first. The informativeness of an example is often measured by either the uncertainty of the model about that example, the expected model change (or reduction in model variance) after training on the example, how representative the example is about other unlabeled examples, or a combination of these measures, see~\cite{settles2012active} for a comprehensive survey. After the example is suggested and its label is obtained, the model retrains and suggests the next example. In practice, this scenario is often unrealistic: in order to be able to implement such a loop in a production system, the AL strategy has to be closely integrated with the annotation system. This, in turns, limits the owner of the dataset to only specific annotation systems. Moreover, many Machine Learning algorithms (such as Decision Trees) don't re-train sequentially, one example at a time, or re-training does not provide a statistically significant impact on the model, as is the case for many Deep Learning models. These models should be retrained on the whole labeled dataset once an additional label is obtained. With the hours-long or even days-long delays for training modern Deep Convolutional Neural Networks (Deep CNNs), this is an unacceptable scenario for most practical systems. 
	
	Instead, it is often suggested that top $B$ informative examples are sent for annotation at once. This is a reasonable heuristic, but might lead to the situation where most or all of the examples are too similar to each other, as we can expect that the model is uncertain about similar examples. This is especially true for the datasets with a lot of redundant examples. Thus a more diverse set might benefit the model more.
	
	In this paper, we suggest an algorithm which incorporates both informativeness and diversity to select a mini-batch of examples which need to be annotated next. Our algorithm can be applied with any learning model, and selected informativeness measure. Diversity is based on the distance of the examples to each other, and the distance metric can be selected according to the user's preferences.
	
	
	The following are contributions of this paper: we connect the problem of diverse selection to the Facility Location problem~\cite{mladenovic2007p}, and propose to solve it with the K-means clustering algorithm in a more scalable way than the previously studied approaches. We empirically demonstrate the benefits of this selection procedure for training simple generalized linear models, multilayer perceptrons, and Deep Convolutional Neural Networks. We suggest a way to incorporate both diversity and informativeness of the examples into account, and empirically demonstrate the benefits of this combination. Finally, we choose to use margin-based uncertainty sampling (see page 14 of~\cite{settles2012active}) as the informativeness measure for the examples, and demonstrate that this informativeness measure works better than random selection in all of the cases, unlike the entropy-based uncertainty sampling used in most other studies.
	
	\section{Related Work}
	Four areas of Active Learning are most related to this work. In one, the selection procedure is made to select both informative and representative samples, see e.g.~\cite{huang2010active},~\cite{hoi2009semisupervised}. These procedures often optimize a specialized objective function and result in using specialized learning algorithms. They do not take into account diversity of the examples in a mini-batch, however, thus falling into the same trap as the informativeness-only based procedures. These methods can be used in the scheme proposed in this paper by providing every example with informativeness score which also incorporates representativeness.
	
	Another approach is to use semi-supervised clustering to select further data points which need to be labeled, see e.g.~\cite{dasgupta2008hierarchical}. The algorithm tries to uncover the structure in the data, through iteratively refining clusters by sampling labels in potentially impure clusters. Whereas related to our work in that clustering of the data is performed, this approach tries to find good clustering of the whole dataset, essentially building a separate classifier from the learning model which is being trained. This approach potentially wastes efforts on building clusters which are already clear to the learning model.
	
	Alternatively, one can consider the scheme where labels of unlabeled examples are hypothesized (for example, by using the most probable label), and then the examples in a mini-batch are selected sequentially after retraining the classifier on the most informative example so far. However, considering all possible label assignments is not computationally tractable, thus most probable labels according to the classifier built so far can be taken instead, or some other heuristic (see e.g.~\cite{sourati2016classification}). This approach significantly narrows the space of possible label assignments and can diverge a lot from the actual labeling. Moreover, the classifier needs to be retrained after every example is added in a greedy manner, which might not be suitable for many learning methods due to time taken to retrain the model.
	
	Most related to this work are papers of~\cite{wei2015submodularity},~\cite{hoi2006batch}, and~\cite{sener2017active}. They incorporate diversity in the mini-batch through formulating a submodular function on the distances between the examples, and selecting a subset of unlabeled examples which optimizes the submodular objective. Our work differs from this in the following ways. First, we use clustering algorithms to find a solution, rather than formulating it as a submodular optimization problem, which leads to better scalability. Second, we use the informativeness score explicitly in the optimization objective by combining it with diversity objective. Third, we demonstrate the efficiency of using margin-based uncertainty.
	
	\section{Problem Setup}
	Let's define by $\mathcal{X}$ the set of all examples $x \in \mathcal{X}$, by $\mathcal{X}^L \subseteq \mathcal{X}$ the set of already labeled examples, and by $\mathcal{X}^U=\mathcal{X}\setminus\mathcal{X}^L$ the set of size $N$ of all unlabeled examples. At every step of Active Learning, we need to select $B$ examples from $\mathcal{X}^U$ for manual annotation and further training of the learning model. We say that the selected examples make the set $\mathcal{S} \subseteq \mathcal{X}^U$, $B = |\mathcal{S}| \le N$. In order to increase the diversity of the selected examples, we formulate the following minimization objective to construct $\mathcal{S}$:
	\begin{equation}\label{eq:medoids}
	f(\mathcal{S}) = \sum_{x_i \in \mathcal{X}^U} \min_{x_j \in \mathcal{S}} d(x_i, x_j)
	\end{equation}
	over $\mathcal{S}$, where $d(x_i, x_j)$ is a distance metric between examples $x_i$ and $x_j$. 
	
	This is a formulation of the Facility Location problem as known in the optimization literature~\cite{wolf2011facility}. Finding the optimal solution is an NP-hard task, thus approximate algorithms should be applied. \cite{wei2015submodularity} shows that the problem can be expressed as a constrained submodular maximization problem. Even though greedy optimization algorithm has provable worst-case guarantees up to some $\varepsilon$,~\cite{fisher1978analysis}, the values of $\varepsilon$ are rarely small so a suboptimal solution is often found~\cite{mladenovic2007p}. 
	
	We can reformulate Equation~(\ref{eq:medoids}) as a $K$-medoids problem (see Chapter 9 of~\cite{bishop2006pattern}). 
	For the purposes of keeping the algorithm more scalable, we suggest using K-means clustering algorithm~\cite{macqueen1967some} to solve it with the Euclidean distance metric. K-means has linear complexity $O(NBI)$ over size of the unlabeled set $N$, batch size $B$, and number of iterations $I$. This greatly improves over $O(N^2)$ computations needed to optimize the submodular function. Even if all distances are precomputed before the Active Learning process starts and approximations to the greedy algorithm are applied, the distance matrix calculation has higher complexity than that of K-means, and the matrix will need to be placed in memory to have reasonable computation speeds. Moreover, K-means is an extremely well studied clustering algorithm and various further improvements exist to scale it to extremely large datasets (see e.g.~\cite{bahmani2012scalable}).
	
	K-means clustering attempts to minimize 
	\begin{equation}
	\sum_{x_i \in \mathcal{X}^U} \sum_{k} z_{i,k}\|x_i-\mu_k\|^2
	\end{equation} 
	by finding cluster centers $\mu_k$ and cluster assignment $z: z_{i,j} \in \{0,1\}$, where $z_{i,k}=1$ and $z_{i,j}=0$ for $j \ne k$. Then, in order to select examples to be labeled, we choose those which are closest to cluster centers.
	
	Assume we are also given informativeness scores $s_i \in [0,1]$ for every example (such that not all of them are $0$). Informativeness can be given by any other Active Learning  algorithm, including those designed for sequential selection: uncertainty sampling~\cite{settles2012active}, or variants of Mutual Information-based selection~\cite{mackay1992information}. We propose to modify the objective function to incorporate the informativeness:
	\begin{equation}\label{eq:weighted}
	\sum_{x_i \in \mathcal{X}^U} z_{i,k}s_i\|x_i-\mu_k\|^2 \to \min,
	\end{equation} 
	and solve it with weighted K-means clustering algorithm.
	During the iterative procedure used to find cluster centers, the optimization for $z_{i,k}$ stays the same: points are always assigned to their closest cluster centers. By taking the derivative of the objective relative to $\mu_k$, it is easy to show that 
	\begin{equation*}
	\mu_k = \frac{\sum_i z_{i,k} s_i x_i}{\sum_i z_{i,k} s_i}.
	\end{equation*}
	
	
	\section{Experimental Results}
	We evaluate our algorithm, formalized as Algorithm~\ref{alg:DBAL}, on the problem of multi-class classification for text and image datasets, using generalized linear models, multi-layer perceptrons, and Deep CNN models. For the informativeness measure, we use margin-based uncertainty in all experiments (see~\cite{settles2012active}): informativeness $s_i$ of an example $x_i \in \mathcal{X}^U$ is defined as
	\begin{equation*}
	s_i = P(\hat{y}_{1,i})-P(\hat{y}_{2,i}),
	\end{equation*} 
	where $P(\hat{y}_{1,i})$ is the predicted probability of the most confident class, and $P(\hat{y}_{2,i})$ is the probability of the second most confident class. This measure of uncertainty performed significantly better in all our experiments, whereas other more popular measures such as entropy-based measure or the least-confident measure often performed worse than a random selector.
	
	At each step, we select $k$ examples for additional model training. We used $k=100$ for all datasets except CIFAR-10, and $k=1000$ for CIFAR-10. In order to further improve scalability of the approach and have a fair comparison with the benchmarks, we do not cluster all the unlabeled examples, but first prefilter them by selecting $\beta k \ge k$ most informative ones. The results are not sensitive to the choice of $\beta$ within certain limits, however, good choice of $\beta$ does depend on the relative size of the dataset and batch size $k$. We found that if the dataset is very large in comparison with $k$, large values of $\beta$ should be used, as diversification plays a big role in selecting among a large set of examples. We used $\beta=10$ to pre-filter 1000 examples (10000 for CIFAR-10 dataset). If the dataset is quickly exhausted however (e.g. selection is planned to be done to 60-80\% of the data), smaller values of $\beta$ lead to better results.
	
	In order to explore various ways of being completely independent on the choice of $\beta$, we have tried different strategies of selecting examples from the clusters. In our experiments with one of the datasets, selecting most informative example from every cluster instead of selecting that closest to its cluster center, lead to the same results as weighted clustering even when clustering is not weighted. Another approach we tried is selecting $k$ examples with largest aggregate scores of similarity to the cluster center and the informativeness. Specifically, distances to cluster centers $\tilde d$ were normalized to 0-1 across all unlabeled examples, and the score of an example $x_i$ with informativeness $s_i$ was computed as $1-\tilde d(x_i, x_c) + s_i$, where $x_c$ is the cluster center where $x_i$ lies. The results of this approach were worse than for other methods, possibly due to the fact that some clusters were not presented in a batch at all as other clusters were much more dense and contributed more than one example.
	
	The first batch is always selected randomly for the comparison to be consistent among different methods. However, we found that if the first batch is selected with our diversity-based approach, the results for all datasets except CIFAR are better than the results for random selection, for the first few batches. We demonstrate this below for MNIST dataset.
	
	We compare the following selection algorithms. Random selection of a batch of examples (denoted Random in the figures), uncertainty sampling of top $k$ uncertain examples (denoted Uncertainty), K-means clustering of prefiltered $\beta k$ examples with $k$ clusters (denoted Cluster($\beta$)), weighted K-means clustering of prefiltered $\beta k$ examples (denoted WCluster($\beta$)), optimization of~(\ref{eq:medoids}) using $\varepsilon$-greedy submodular optimization after prefiltering $\beta k$ examples (denoted Submodular($\beta$)), and $FASS$-framework from~\cite{wei2015submodularity} using Nearest-Neighbor-based $f_{NN}$ and $\varepsilon$-greedy submodular optimization after prefiltering $\beta k$ examples (denoted FASS($\beta$)). 
	
	As the first batch is selected randomly, we repeat the Active Learning procedure multiple times, and average the results from multiple repetitions. All figures have the $90$-th confidence interval band around the mean. Horizontal axis is number of examples labeled so far, and vertical axis is the test accuracy of the classifier trained on those examples.

	\begin{algorithm}[tb]
		\caption{Diverse mini-Batch Active Learning (DBAL)}
		\label{alg:DBAL}
		\begin{algorithmic}
			\State {\bfseries Input:} dataset of examples $x_i$, budget $B$, batch-size $k$, pre-filter factor $\beta$
			\State Select first $k$ examples randomly, obtain labels for these examples
			\Repeat
			\State Train classifier on all the examples selected so far
			\State Get informativeness for every unlabeled example
			\State Prefilter to top $\beta k$ informative examples
			\State Cluster $\beta k$ examples to $k$ clusters with (weighted) K-means
			\State Select $k$ different examples closest to the cluster centers, obtain labels for these examples
			\Until{Budget $B$ is exhausted}
		\end{algorithmic}
	\end{algorithm}
	
	\subsection{Browse Node UK Appliances}
	This Amazon dataset contains product titles and descriptions (which we use as text features, concatenating both) for 9K products which were sold in 2015 at the Amazon UK marketplace. The label is the category of the product. The dataset has products from 24 categories with unbalanced classes: the largest category has 27\% of the examples, and the smallest has 0.07\%.
	
\begin{figure}[ht]
		\vskip 0.2in
		\begin{center}
			\centerline{\includegraphics[width=0.9\columnwidth]{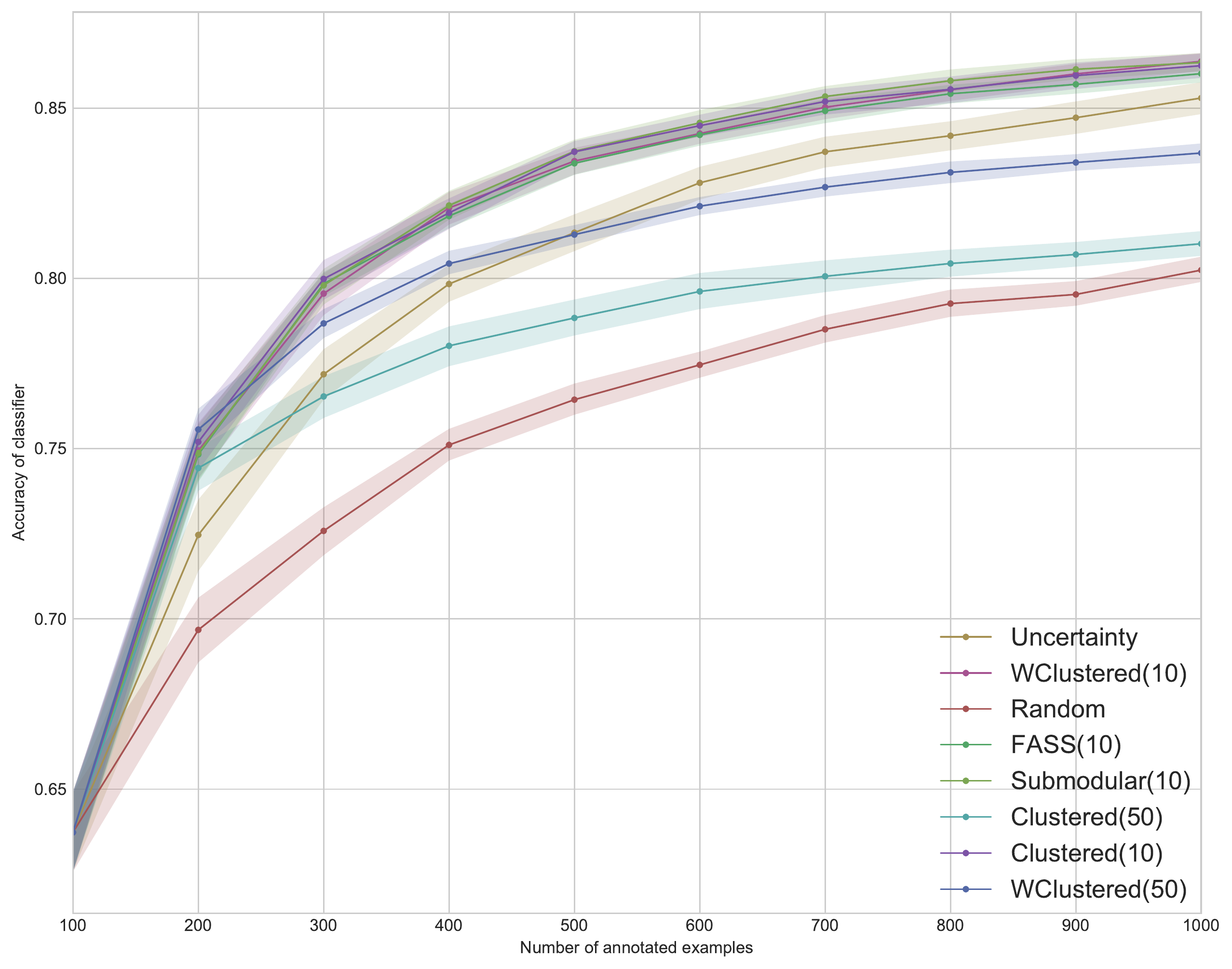}}
			\caption{Accuracy on Browse Nodes UK dataset}
			\label{fig:bnuk}
		\end{center}
		\vskip -0.2in
	\end{figure}

	We build a logistic regression classifier for this dataset. To transform text features, we used the pipeline of \lstinline|CountVectorizer(ngram_range=(1, 2))|,  \lstinline|TfidfTransformer(use_idf=True)|, and \lstinline|Normalizer()| functions from Scikit Learn package~\footnote{\url{http://scikit-learn.org/}}.
	Before performing Active Learning, we randomly split the dataset into train and test sets, with 70/30 split. The results presented in Figure~\ref{fig:bnuk} are averaged among 16 random splits.
	
	We can see that for the same value of the parameter $\beta=10$, all diversity-based methods perform similarly to each other, and significantly outperform the baseline. However, the Clustered method is significantly faster than the submodularity-based methods (with the same hardware and parallelization setup, experiments with FASS(10) finished in about 4700 seconds, whereas experiments with Clustered(10) finished in only 10 seconds). An important observation is that for large value of $\beta=50$, diversity-based methods start performing worse than the benchmark of Uncertainty sampling after the classifier is trained on about 300 examples. This can be explained by the fact that the classifier becomes sufficiently good for this dataset, and a large informative portion of the dataset has been exhausted already, so the selection methods should rely more on the informativeness. Interestingly, the weighted clustering approach becomes worse than Uncertainty sampling only after the classifier is trained on about 500 examples, indicating the usefulness of weighting examples for this dataset.
	
	\subsection{20 Newsgroups}
\begin{figure}[ht]
		\vskip 0.2in
		\begin{center}
			\centerline{\includegraphics[width=0.9\columnwidth]{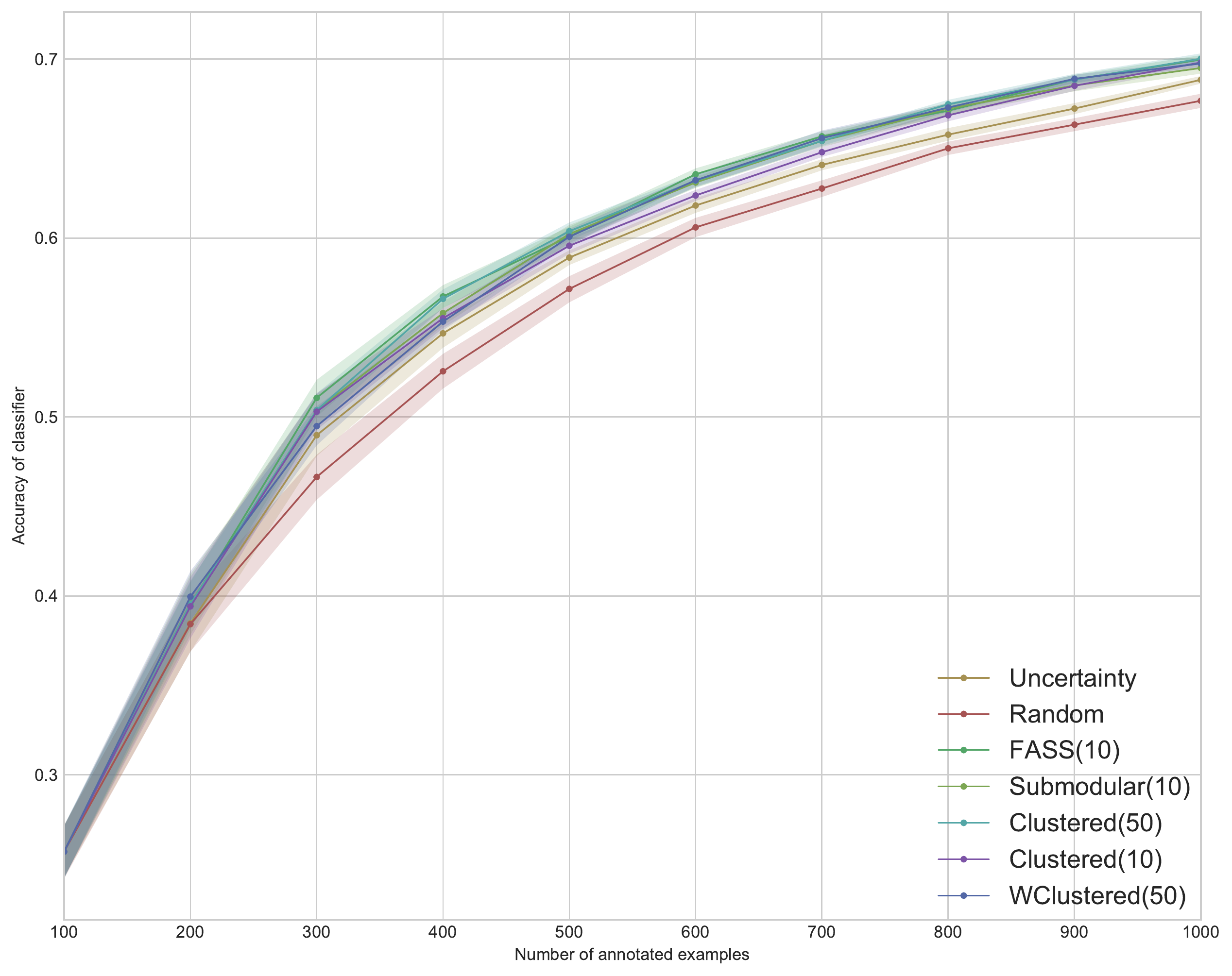}}
			\caption{Accuracy on 20 Newsgroups Dataset}
			\label{fig:20ng}
		\end{center}
		\vskip -0.2in
	\end{figure}

	20 Newsgroups dataset\footnote{Obtained with \lstinline{fetch_20newsgroups} command from Scikit Learn package} contains 11314 train and 7532 test articles sent to one of the 20 UseNet discussion groups. The goal is to classify which of the newsgroups the article was sent to. To transform text features, we use the pipeline of \lstinline|CountVectorizer(ngram_range=(1, 2), max_features=250000)| and \lstinline|TfidfTransformer(use_idf=True)| functions from Scikit Learn, and build a multinomial logistic regression classifier. 
	Figure~\ref{fig:20ng} presents the results for that dataset, where the curves present the mean accuracy among 16 runs. We omit WClustered(10) from this figure as it performs similarly to Clustered(10).
	
	We can see that for this dataset, all the methods which incorporate diversity perform slightly better than the baseline of Uncertainty sampling. For the same value of pre-filtering parameter $\beta=10$, K-means clustering performs on the lower range of the confidence interval of the submodularity-based methods. However, clustering with higher value of $\beta=50$, performs comparable, and at the same time is still significantly faster than the submodular methods for $\beta=10$.
	
	\subsection{MNIST}
	MNIST dataset\footnote{Obtained with \lstinline{mxnet.gluon.data.vision.MNIST} command from MXNet package} has 60,000 training and 10,000 test examples. Each example is an image of a handwritten digit. The task is to identify the digit by its image. The training and test data are both almost evenly divided among 10 different classes. For this dataset, we used a simple multilayer perceptron as a classifier, with 2 dense hidden layers with 128 and 64 units and ReLU activation, and an output layer~\footnote{\url{https://github.com/gluon-api/gluon-api/blob/master/tutorials/mnist-gluon-example.ipynb}}. We present average among 16 runs.
	
\begin{figure}[ht]
		\vskip 0.2in
		\begin{center}
			\centerline{\includegraphics[width=0.9\columnwidth]{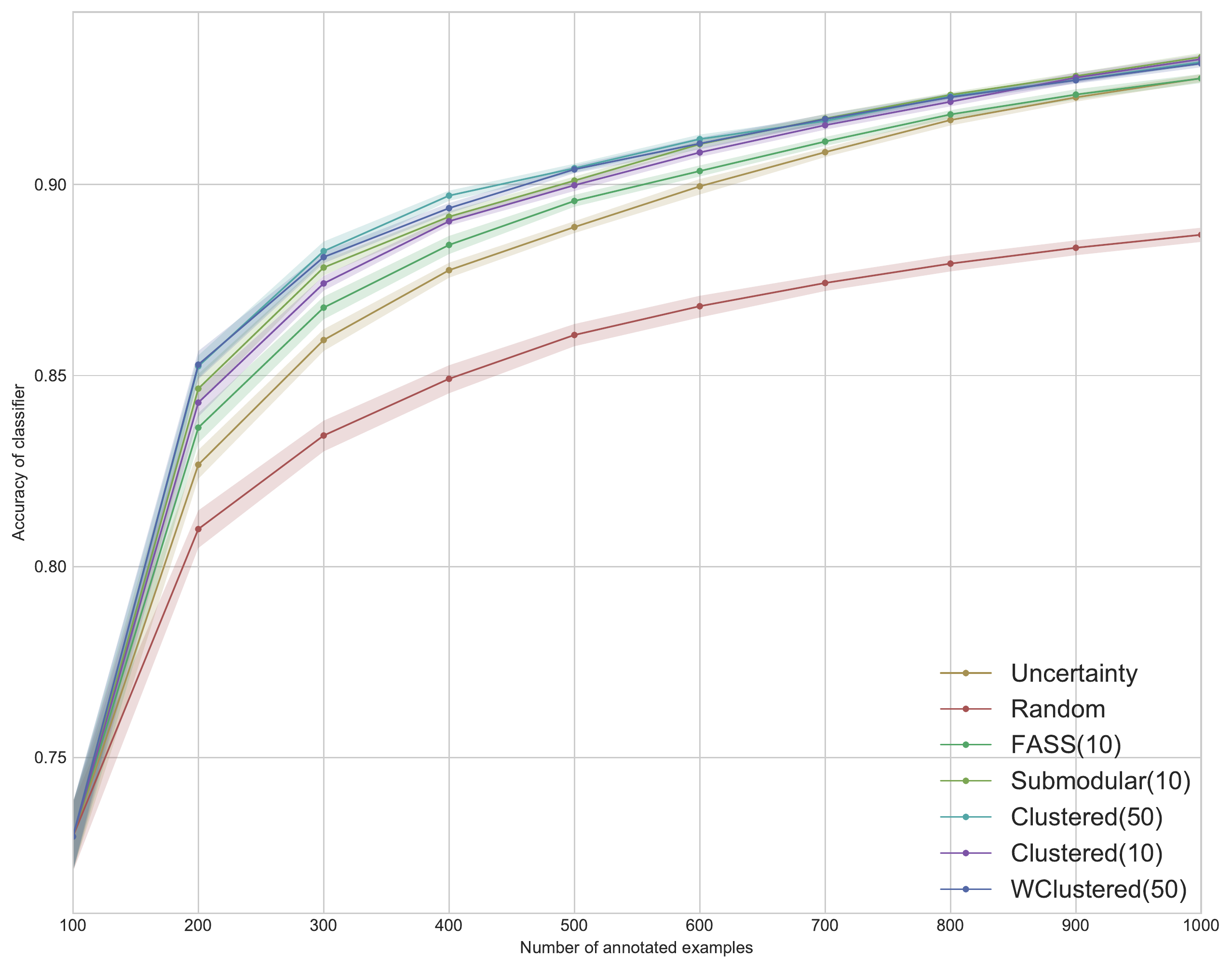}}
			\caption{Accuracy on MNIST dataset}
			\label{fig:mnist}
		\end{center}
		\vskip -0.2in
	\end{figure}

	From Figure~\ref{fig:mnist}, we can see that all diversity-based methods significantly outperform the baseline of uncertainty sampling. Both weighted and unweighted clustering methods with large prefiltering $\beta=50$ outperform diversity-based methods with smaller prefiltering of $\beta=10$, and for the same value of $\beta$, our proposed method performs as well as Submodular and better than FASS.
	
\begin{figure}[ht]
		\vskip 0.2in
		\begin{center}
			\centerline{\includegraphics[width=0.9\columnwidth]{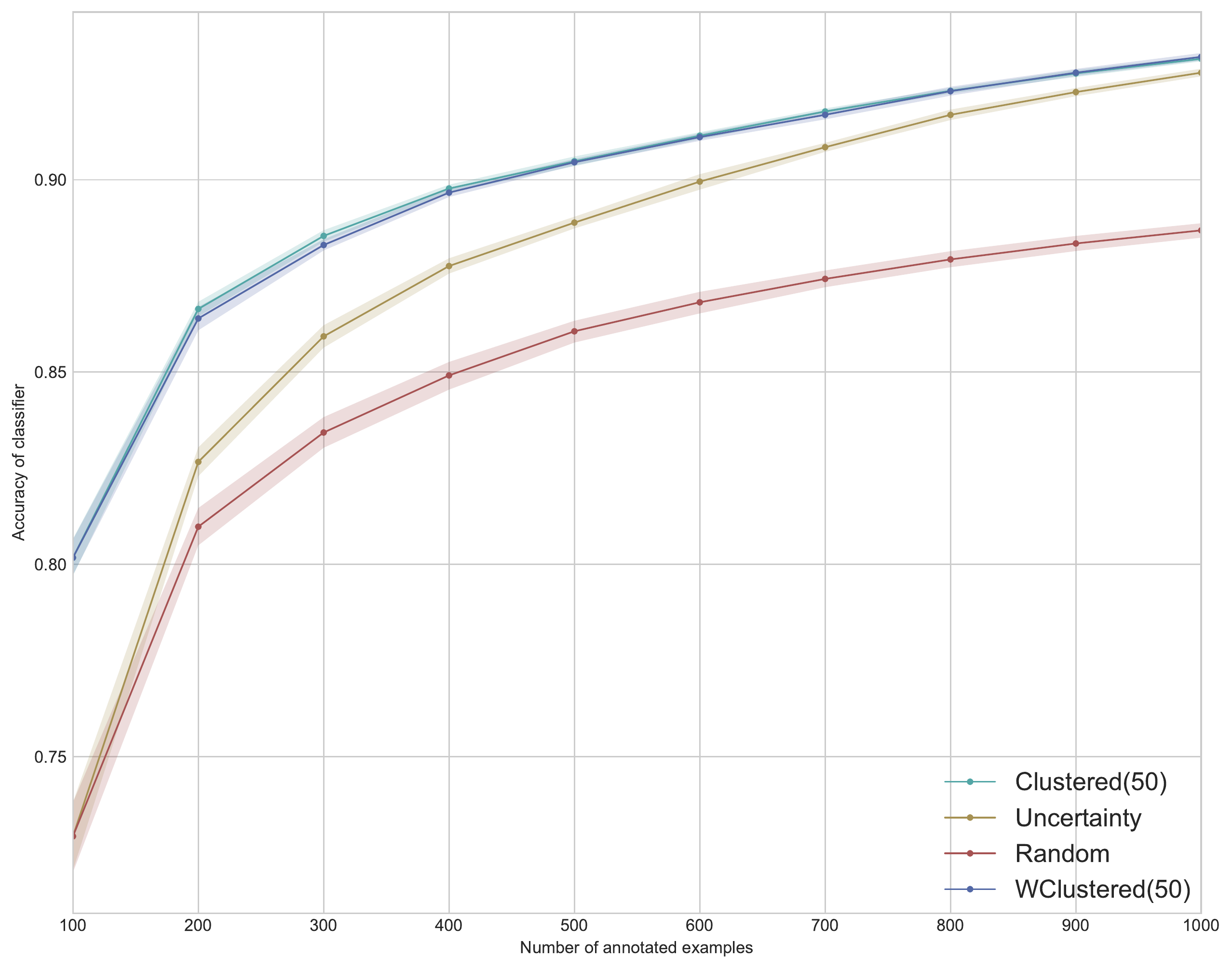}}
			\caption{Accuracy on MNIST dataset, including the methods which select the first batch using clustering}
			\label{fig:mnist_clstart}
		\end{center}
		\vskip -0.2in
	\end{figure}

	To demonstrate the potential of clustering from the beginning of the Active Learning process, we present Figure~\ref{fig:mnist_clstart} where we used K-means to select first batch of examples, rather than selecting them randomly as in other experiments. We can see that accuracy is higher in the first two steps than the accuracy of all the methods presented on Figure~\ref{fig:mnist}.
	
	\subsection{CIFAR-10}
	CIFAR-10 dataset~\footnote{Obtained with \lstinline{mxnet.gluon.data.vision.CIFAR10} command from MXNet package} has 50,000 training and 10,000 test examples. Each example is a 32x32 color image of an object or an animal, with 10 classes in total. The images are evenly divided among the classes. For this dataset, simple models do not progress far in training, and we used Resnet~\cite{he2016identity} Deep Convolutional Neural Network~\footnote{The model without pretrained weights is obtained with \lstinline|mxnet.gluon.model_zoo.vision.resnet34_v2| command}. This model requires more data for learning meaningful weights, thus we increased the batch size to $k=1000$ and budget to $B=10000$. With this data size, running submodularity-based methods is prohibitively expensive, thus we only present results for the clustering-based methods. Notice that we do not perform any random data augmentation (such as horizontal flip, etc.), which is often used to achieve much higher accuracy numbers on CIFAR-10 dataset.
	
	For this dataset, we leveraged the capabilities of CNNs to provide compact data representation. We featurized the examples by passing them through the CNN trained so far, and used the outputs of the last pre-final layer as the vectors for clustering.
	The results are averaged among 8 runs.
	
\begin{figure}[ht]
		\vskip 0.2in
		\begin{center}
			\centerline{\includegraphics[width=0.9\columnwidth]{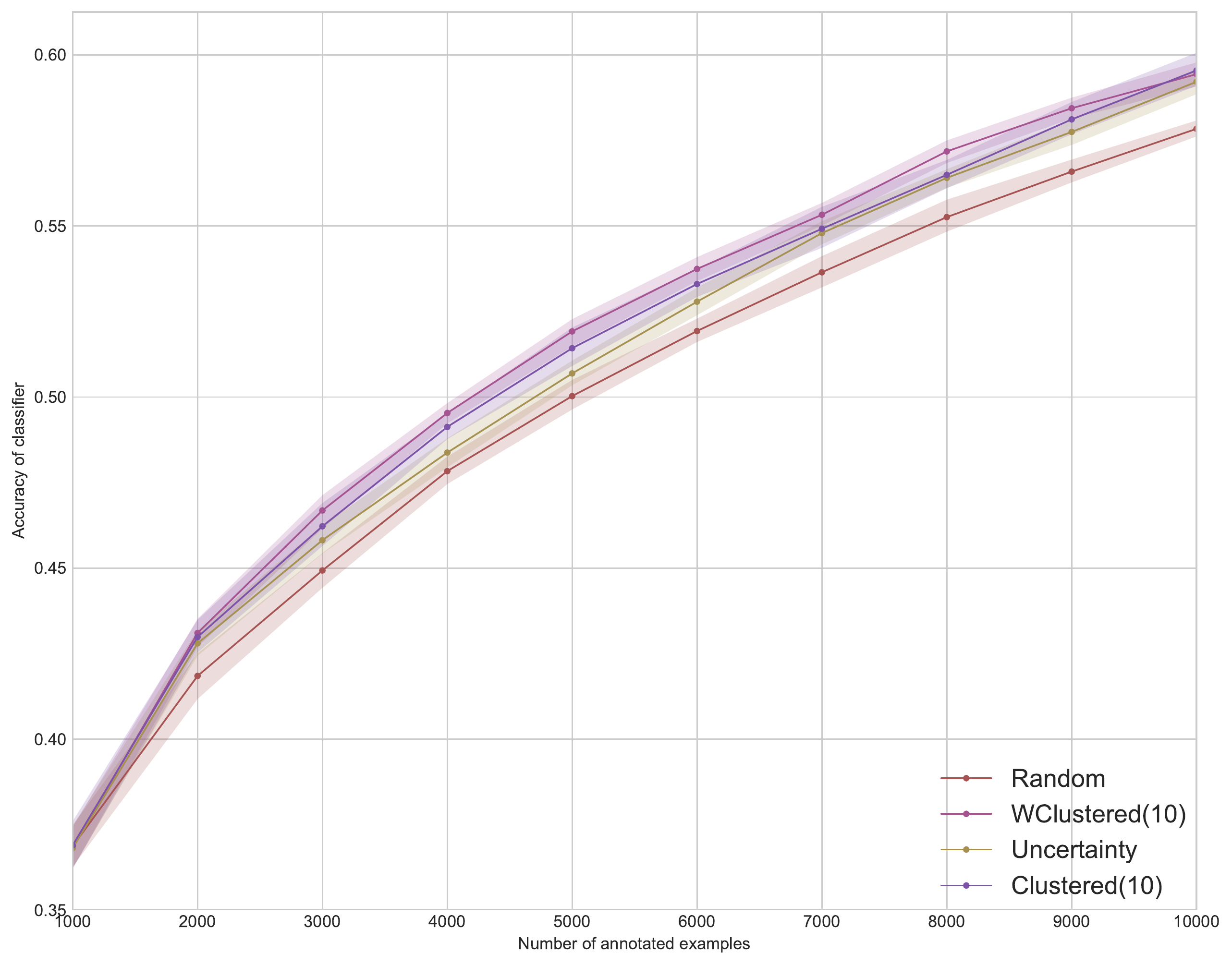}}
			\caption{Accuracy on CIFAR-10 dataset}
			\label{fig:cifar10}
		\end{center}
		\vskip -0.2in
	\end{figure}

	Figure~\ref{fig:cifar10} shows that diversity-based selection slightly outperforms the baseline of uncertainty sampling, and weighted clustering in turn outperforms the non-weighted version.
	
	\section{Discussion}
	In this paper, we proposed a scalable approach to increase diversity in mini-batch Active Learning, and linked the approach to Facility Location. We have experimented with the datasets of various sizes and nature, using models of various complexity from generalized linear models, to Deep CNNs. All experiments show that diversity-enhancing approaches slightly or significantly outperform the strong baseline of uncertainty sampling. We also show that the proposed approach is achieving the performance comparable to the previously published techniques, but is intrinsically more scalable. Moreover, in all the experiments, we demonstrate the efficiency of the selected baseline, even though this baseline is usually not chosen in other studies.
	
	For further research directions, it is interesting to study methods for further reducing dependency on the selection of the pre-filtering parameter, as well as to test scalable analogues of the K-means approach designed for non-euclidean distances.

\bibliography{../bibliography}
\bibliographystyle{plain}
\end{document}